# Cyber-physical attack detection approaches in the wide-area measurement system

Zakaria El Mrabet, Prakash Ranganathan
University of North Dakota, Grand Forks, ND, USA
Email: zakaria.elmrabet2@gmail.com

*Abstract*— **Improving situational awareness within the Smart Grid is an effective preventive measure to detect potential anomalies, and avoid their adverse consequences. A wide area measurement system (WAMS) based Phasor Measurement Unit (PMU) can assist in improving visibility within the grid by collecting relevant measurements including magnitude, phase angle, frequency, voltage, and current, with a precise GPS-based timestamp. So far, there are more than 2500 PMUs deployed in the North American power grid, where each of which generates between 30 and 60 samples/s, roughly 1.5 TB/month of streamed data. Thus, processing such large PMU streams requires expensive computational resources and faster algorithms. This paper provides state-of-the-art cyber and physical anomalies in the WAMS and analyses their impact on the integrity and availability of the system. Next, it investigates the current anomaly detection approach in historical and streaming PMU data, including conventional machine learning and hybrid approaches, and assesses their advantages and main drawbacks. Lastly, it sheds light on the importance of handling missing PMU data and concept drifts.**

## I. Introduction

In 2015, the Ukraine power system has experienced an outage for several hours impacting 225 000 utility customers. One year later, another cyber-attack was reported on a power control system in another city in Ukraine shutting down substations that controlled 200 megawatts of capacity [1]. These two succeeding events illustrate the threats posed by cyber-attacks and their malicious ability in disturbing the power system and making a negative social-economic impact. One of the reasons behind these cyber-attacks and several major blackouts, especially in North America, is the lack of system awareness and reliable measurements [2].

Smart Grid harnesses the power of information technology to deliver intelligently the power to the end customer and meet wisely the environmental requirements. It allows two-way communications between the end-user and the utility and facilitate the integration of renewable technology. Smart Grid is a distributed, interconnected, and interdependent system. Thus, an event in one part of the grid could impact operations elsewhere. It could also go beyond the grid to affect other system that relies on stable power. Thus, time-synchronized measurements is needed to obtain an accurate real-time estimation of the state of the grid and react quickly to the suspicious events [3].

## II. Wide Area Measurement System

The WAMS based PMUs provides a situational awareness, operation and reliability of the power system network [4]. It plays a major role in the Smart Grid network by providing a near time-synchronized measurement which enables the grid operator to analyze the operating status of the power grid. Such a feature is achieved through the utilization of the synchrophasor which synchronizes the critical measurements received from different grid components widely distributed in the power grid with a precise clock through the Global Positioning System (GPS). The PMU is regarded as the nervous system of the power grid as it is capable of providing time-synchronized measurements of the grid status, which enable the grid operators to react quickly to accidental and unexpected events.

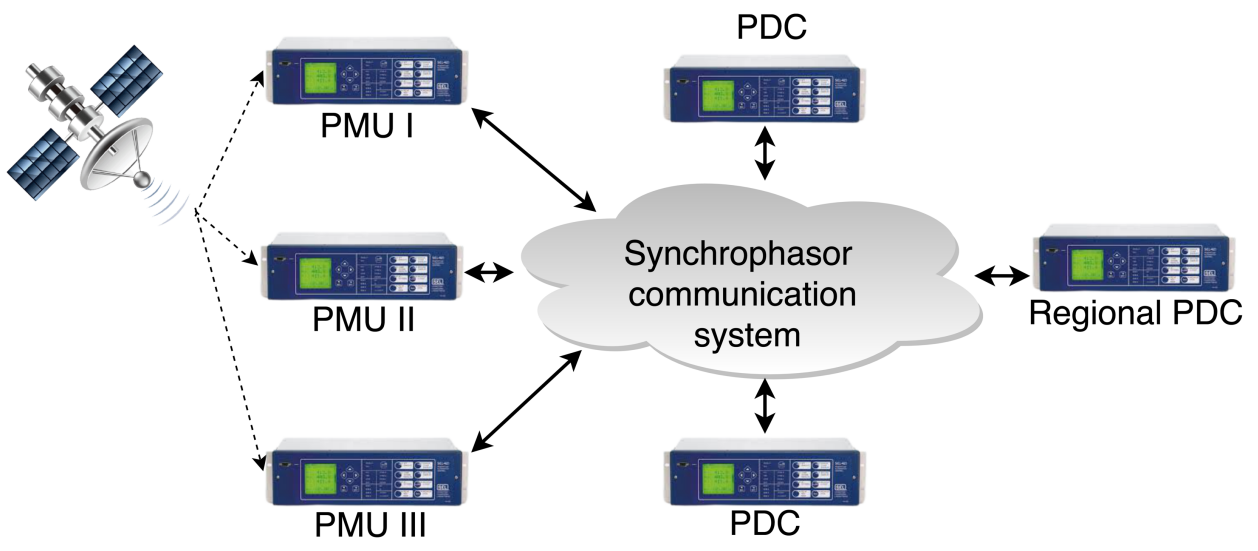

Figure 1 A typical PMU-based WAMS architecture

Since the time-synchronized measurements produced by the PMUs are sent to Phasor Data Concentrator (PDC) or a control center, it is important to protect all the components in the synchrophasor communication system. this system is called the PMU based Wide Area Measurement System (WAMS). A typical WAMS is composed of PMUs, and local and regional PDC, As illustrated in Figure 1.

The PMU receives the three-phase voltages and currents coming from the power system network in the form of an analog signal. This signal is filtered through an anti-aliasing module to limit the bandwidth of the incoming signal and then converted into a digital signal via the (A/D) converter module. The digital signal is fed to the Central Processing Unit (CPU) to compute the phase and magnitude of the signal using the Discrete Fourier Transformer in conjunction with the timed GPS signal. The output signal is time-synchronized using a sampling clock that is phase-locked to the one-pulse-per-second provided by the GPS receiver. The pulse signals from the satellite are phase-locked with the sampling clock through the Phase Locked Oscillator (PLO) module. PLO divides the one pulse per second signal from GPS into the required number of pulses per second for sampling [5][6]. The time-stamped signal is transmitted to the local PDC. Then, the measurements from several local PDC are reported to the regional PDC. As shown in Figures 1 and 2.

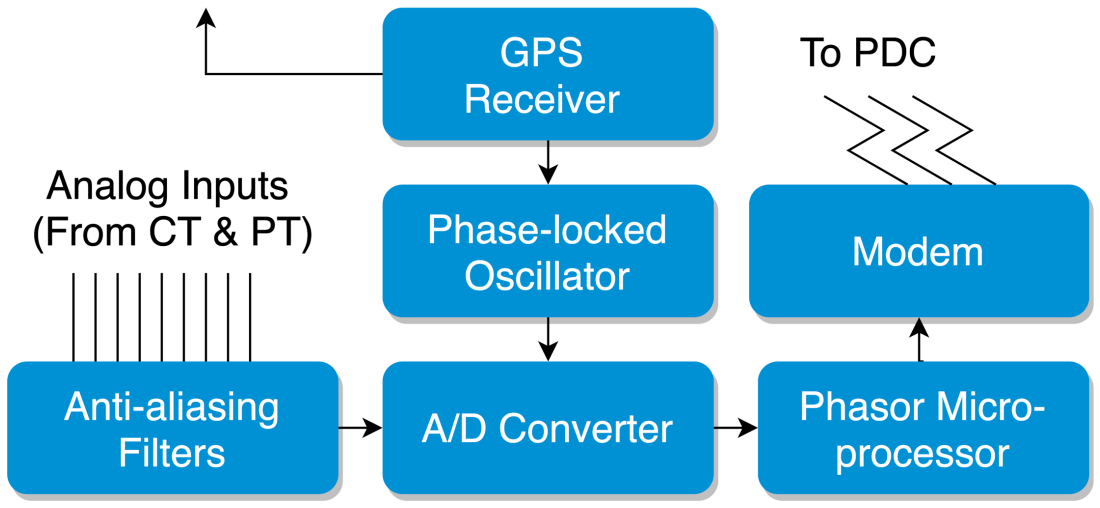

Figure 2. Phasor measurement unit components

The communication between the PMUs and PDC is defined through various communications technology including Power Line Communication (PLC), Optical Fiber communication, Microwave communication, Cellular communication, and Satellite communication. The end-to-end communication between the PDC and the PMU is also based on the IEEE C37.118.2-2011 standard which defines the format of the message for exchanging data between the PMU and the PDC or between PDCs. There are four types of message types: data, configuration, header, and command frames. The header, configuration, and data frames are sent from the PMU to the PDC while the command frames are sent from the PDC to the PMU. The header frame provides information about the PMU and helps the PDC identify the PMU. In the configuration frame, there are three types: CFG-1 which provides information about the capacity of the PMU, such as reporting rates and noise suppression, and CFG-2 and CFG-3 which provides information about the currently reported measurements. The data frame includes the real-time synchrophasor data measured by the PMU including the current, frequency and voltage, and amplitudes.

Figure 3 shows a typical real-time communication between a PMU and a PDC through the IEEE C37.118.2-2011 standard. The PDC initiates the communication by sending a command frame to turn on the transmission and sends a request for a configuration frame (CFG). The PMU responds to this request, and then PDC makes another request for synchrophasor data. The PMU sends back the requested data which is included in a data frame format. In the data frame parquets, there are various fields which serve in initiating and terminating the communications. For instance, the SYNC field marks the beginning of the frame, a Frame Size field provides the length of the entire frame, an IDCODE field is used as an ID to identify the frame, a SOC field provides information about the time stamp, A FRACSEC field which presents the time quality or time of measurements for data frames, and then a Cyclic Redundancy Check (CRC) which marks the end of the data frame [7] [8].

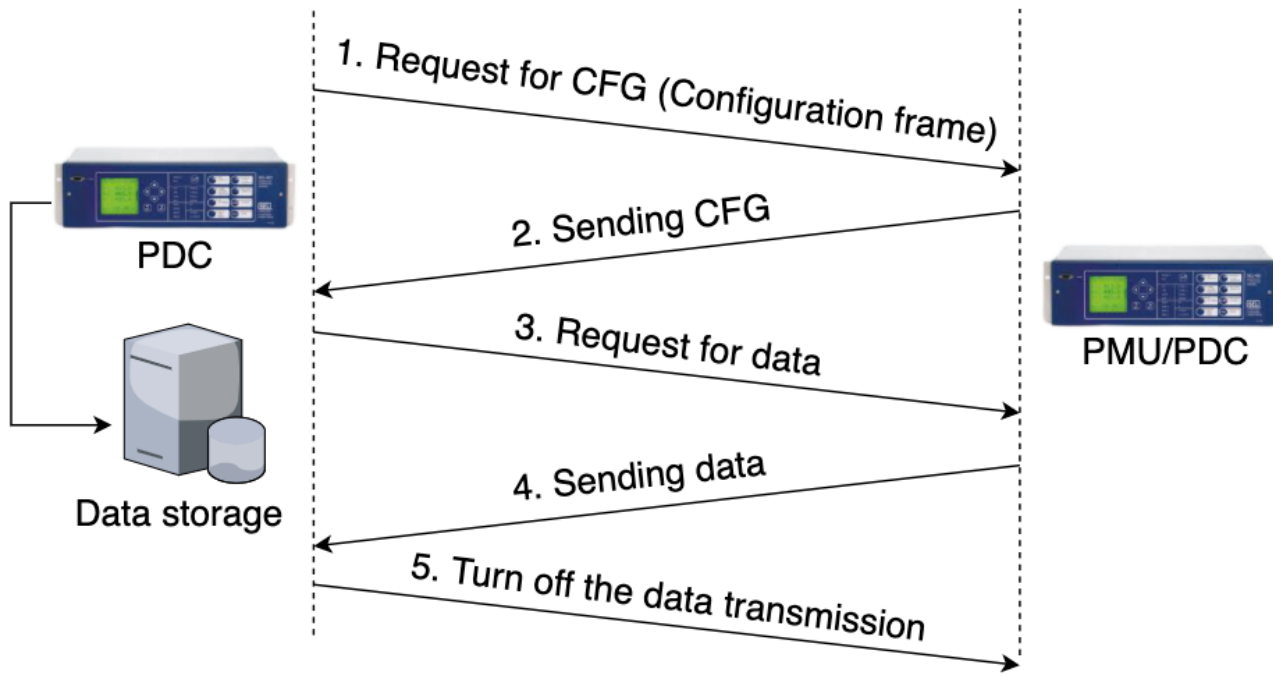

Figure 3. PMU-PDC communication via the IEEE C37.118.2 protocol

## III. Cyber-physical Anomalies on PMU

AWAMS system incorporates computation and networking with physical proves and relies on PMUs to monitor the power system, creating complex feedback loops between the physical and cyber worlds [9]. The cyber and physical systems are interconnected via information technology (IT) and operational technology (OT). IT refers to the application of networks for storing and transmitting data; while OT refers to monitoring and controlling physical devices deployed on the field [10].

Broadly speaking, anomalies correspond to patterns and abnormal behavior in data. These nonconforming patterns are also referred to as outliers, discordant observation, and exceptions [11]. In the WAMS contexts, anomalies refer to all malfunctions either intentionally or unintentionally caused by physical or cyber events. Here, we classified anomalies into two classes: cyber and physical; all anomalous events that are directed at compromising the IT network are considered cyber-anomalous events, while those that target the OT network are termed physical malicious events. The Physical and cyber anomalies target the availability and integrity of the WAMS.

### *A. Physical anomalies*

#### *1) WAMS availability anomalies*

Anomalies that target the WAMS availability include physical faults in power systems. These faults are classified into balanced and unbalanced faults [5], [6]. An unbalanced fault, also known as an asymmetrical fault, is a commonly occurring fault in the power system and can be of a series or shunt type. In the series fault type, the voltage and frequency values increase, while the current level decreases at the faulty phases. In shunt fault, the current level rises, while the frequency and voltage levels decrease at the faulty phases. There are several shunts faults types: single line to ground (SLG), line to line (LL), double line to ground (DLG), and Three-phase to the ground (LLL). An SLG occurs when a transmission line phase touches a neutral wire or the ground; the DLG or LL fault occurs when two or more phases make a connection with the ground; The cause of this fault is mostly due to natural weather events to storms or high winds or fallen trees. Sometimes, the cause can be due to an equipment failure, line connecting the remained phases, or a failing tower. According to [6], [7] the likelihood of occurrence of each fault type is 70% for SLG, 15% for LL, 10% for DL, and 5% for LLL. Although there is a low likelihood of LL fault occurrence, it is considered a severe fault that can cause a rise in fault current magnitude and there by resulting in outages or large damage to grid assets.

#### *2) WAMS Integrity anomalies*

This category includes the anomalies that compromise the WASM integrity by modifying or injecting falsified data. (a.k.a. False Data Injection (FDI)). FDI are designed to alter data stored in the control center or transmitted via communications infrastructure, which ultimately compromises the integrity of the power systems or even their availability. An attack of this type was first introduced by [12] in connection with DC system models, in which the authors assumed that the adversary had knowledge of the power system topology and parameters and could alter meter measurements while remaining consistent with physical laws such as Kirchhoff's circuit law to bypass the bad data detection process [10]. Such an attack could result in inaccurate estimates of the power system state, which is essential for a variety of power system applications, including economic dispatch and contingency analysis, which in turn could result in the state estimator sending incorrect values to the operator and leading to unstable system states, voltage collapses, and eventually physical and economic losses [13]. The authors of [14] examined the implications of a hidden FDI attack at the RTU level on AC state estimation and concluded that it had two negative consequences.

If the data is altered in a way that the state estimation schemes cannot detect as false, there will be an incorrect perception of the observable state of the system, which may lead the grid operator to take actions that endanger the system's security. Despite the detection of an attack, parts of the system may become unobservable, meaning the state estimator is unable to estimate changes in the state values (e.g., voltage magnitudes and voltage angles), placing the transmission grid at risk from a local physical attack. It may already be too late to prevent an outage of a greater part of the system by the time the effects of the physical attack have propagated into the rest of the system in which the state is observable. It has also been concluded that the FDI attacker using a DC model has a greater chance of introducing errors in the measurements and ultimately triggering bad data detection. Thus, the nonlinearity of the power flow equations provides advantages to the system operator to detect this type of FDI attack. However, if the attacker knows the estimated state, they can launch an AC FDI attack without being detected by the AC state estimation. But in reality, it is difficult for an attacker to obtain the same estimated state as the operators. Thus, authors in [15] proposed a sufficient condition for an imperfect and undetectable FDI attack. Authors in [16] proposed the design of blind FDI attacks based on little to no knowledge of the Smart Grid topology which would significantly reduce the attack cost.

Authors in [17] have suggested another FDI attack on the electricity market, using small changes to the price signals to increase the difference between the generated and consumed power. This attack is extended here [18], in which the attacker can inject false pricing data at any time and repeatedly over some time, resulting in over-generation, economic losses, and poor power quality.

### *B. Cyber anomalies*

#### *1) WAMS availability anomalies*

Jamming attack targets mainly the availability of the WAMS where an adversary exploits the shared nature of the wireless network and sends a random or continuous flow of packets to keep the channel busy and then prevents legitimate devices from communicating and exchanging data [19]. Due to its time-critical nature, Smart Grid requires a highly available network to meet the quality of service requirements and such an attack can severely degrade its performance [19]. Keke G. et al. [20] proposed a jamming attack named maximum attacking strategy using spoofing and jamming (MAS-SJ) that targets mainly the wireless Smart Grid network (WSGN). Because WSGN is important for monitoring the Smart Grid along with the PMUs, which play a key component by providing time-synchronized data of power system operating states [20], attacks like MAS-SJ can disturb the operation of the system or even make it unavailable [21].

Denial of Service are other attacks aim to adversely affect the availability of a system. In the context of a Smart Grid, DoS attacks are typically used to prevent the control center from receiving sensor measurements or actuators from receiving control commands using a variety of methods, including jamming the communication channel by transmitting a signal with high transmission power to flood the targeted channel and eventually block it or violating network protocols to increase packet collisions. While the system operator can detect the attack due to the loss of measurement data, they are unable to stop it due to the inability to send control signals to the actuators [53], [54]. Authors in [8] proposed a DoS-based attack known as the puppet attack that targets the AMI network by exploiting a vulnerability in the dynamic source routing protocol, then exhausting the communication network bandwidth, causing a drop of packet delivery of 10% to 20%. The time -delay-switch attack [55] is another DoS attack that introduces a delay in the control system leading to instability in the system

*2) WAMS integrity anomalies*

A study conducted by Daniel P. et al. [3] showed how a GPS spoofer was able to manipulate the PMU readings and cause a plant to trip. Through their experiments, they demonstrate that an attacker can manipulate the PMU time stamp by injecting falsified set of the GPS signal into the antenna of the PMU's time reference receiver. Injecting timing error for a few microseconds was enough for the PMU to violate the maximum phase error allowed by the applicable standards, Such errors can provide false perception of the status of the grid, leading to unnecessary control actions [3]. Therefore, it is important to protect the PMU from such manipulations.

Viruses are programs that are used to infect a specific device or system and compromise its integrity. Worms are self-replicating programs that take advantage of a network to spread, replicate, and infect other devices and systems [21], [22]. Trojan horses are programs that appear to carry out legitimate functions on the target system, but run malicious code in the background [21], [22]. In June 2010, Roel Schouwenberg, an analyst at Kaspersky Lab, detected Stuxnet, the first worm that attacks SCADA systems [23]. Stuxnet, a worm of 500 KB, exploited many zero-day vulnerabilities, which were not yet disclosed by the software owner. It infected at least 14 industrial sites based in Iran, including a uranium-enrichment plant. More than one year later, two more worms that targeted industrial control systems were discovered, Duqu and Flame. Unlike Stuxnet, Duqu was designed to gather and steal information about industrial control systems. Flame, on the other hand, was created to be used in cyber espionage in industrial networks. It has been found in Iran and other Middle East countries [22], [24].

Man-in-the-middle (MITM) are other data integrity attacks where an attacker inserts themselves between two legitimate devices and listens for, performs an injection, or intercepts the traffic between them. Upon connecting to these devices, the attacker retransmits the traffic between them; the two legitimate devices appear to communicate directly, however, they are communicating through a third device [21], [22]. In the Advanced Metering Infrastructure (AMI), a MITM attacker falsifies the actual energy consumption by intercepting the data exchanged between the customer and the utility via the ZigBee protocol. This protocol is widely used for short-rand wireless data transfer, particularly for wireless low-power devices, but it suffers several vulnerabilities [25]. Thus, the attacker can easily compromise this weakest point of the system and then tap on it to launch more advanced attacks from within the network.

## IV. Anomaly Detection Approaches

Several approaches for detecting cyber and physical anomalies have been reported in the literature [26]- [27]. Some authors have proposed conventional approaches, such as impedance-based and traveling wave techniques, while others have proposed machine learning-based approaches, which can be categorized into supervised, unsupervised, semi-supervised, and reinforcement learning. Other authors have proposed hybrid approaches comprising conventional and machine learning approaches. In this section, we will examine all these various approaches and analyze their advantages and limitations.

### *A. Conventional approaches*

Conventional approaches rely on traveling wave [57] and impedance-based methods [35], [59] for detecting an anomaly in the grid. The traveling wave approach needs high-speed data acquisition equipment, a GPS, sensors, and a transient fault recorder to detect the transient waveform for fault location. The location of the fault is computed by tracking" time-tagging the arrival of the traveling wave at each end of the line and comparing against the time difference to the total propagation time of the line with the help of GPS" [57]. This approach has several advantages, as the approach is not impacted by excessive resistance, load variance, reflection, grounding resistance, refraction of the traveling waves ,or series capacitor bank [57]. However, the accuracy of the approach relies on capacitance and line inductance. Unlike the time wave method, the impedance-based approaches [35], [59] are simple and easy to implement, as they require only measurement data that include fault voltages and fault currents collected from the digital fault recorder or relays to compute the impedance. The accuracy of this approach can be affected in the case of a grounded fault, where the fault resistances can reach higher values.

### *B. Machine learning-based approaches*

The machine learning techniques used to detect anomalies in the Smart Grid are categorized as supervised, unsupervised, semi-supervised, and reinforcement learning (RL) [28]-[29]. Supervised learning algorithms build regression or classification models from a set of input features and their corresponding outputs. Although there is a wide range of supervised models, including logistic regression and neural networks as well, they usually fall into one of the following categories: Decision Tree models, Function models, Bayesian models, and Ensemble models [30]. Unsupervised learning is the process of discovering hidden patterns and associations within unlabeled data. The most common unsupervised learning algorithm is the clustering [31]. Semi-supervised learning refers to a combination of supervised learning and unsupervised learning, and it is typically used in cases where training data is insufficient [32]. RL refers to the process of learning directed towards a specific objective in which an agent learns by interacting with an unknown environment, typically in a try-and-error way. In this process, the agent receives feedback from the environment in the form of a reward (or punishment); it then uses this feedback to train itself, as well as to acquire experience and knowledge about the environment [33]. Figure 4 illustrates the various anomaly detection approaches in PMU data. Before describing these various ML categories, we will provide preprocessing ML approaches for preparing unbiased PMU training data and assist the models in converging quickly. The preprocessing includes dimensionality reduction and imputing missing data.

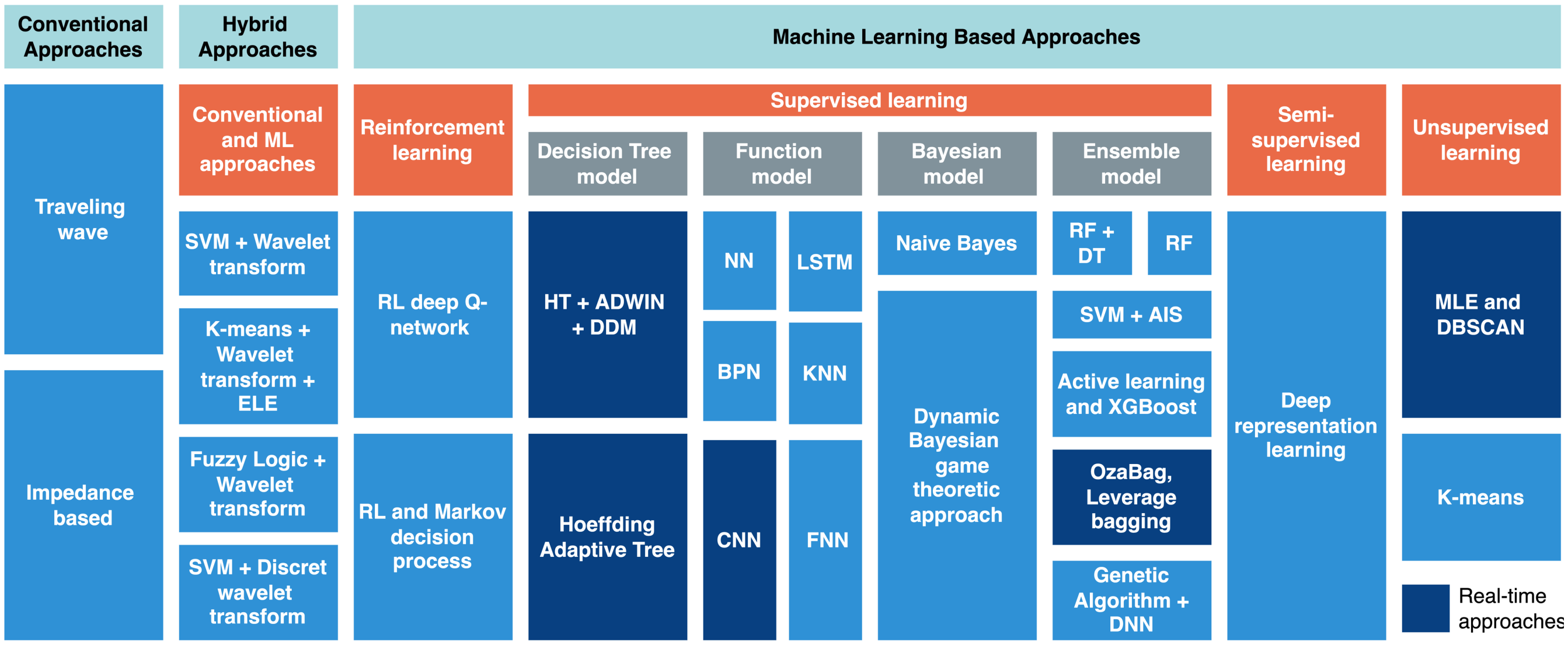

*Figure 4. Taxonomy of anomaly detection approaches in the WAMS*

*1) PMU data preprocessing*

*a) Dimensionally reduction for historical PMU data*

PMUs generate a substantial amount of data exhibiting multi-dimensionality (e.g., time, location, voltage, current, frequency, power flow, and phase angle). It is necessary to reduce data dimensionality to make use of it in real-time decision-making [34]. Xie et al. [35] proposed a PCA-based dimensionality reduction method to keep only the relevant features and detect power system events at an early stage. In [36], PMU data are represented in Spatio-temporal streams. It is highlighted that each PMU measurement contains many categories of variables, such as voltage magnitude, power flow, frequency, etc. However, it is assumed that each type of measurement is independent. This clearly shows the need for higher-order generalizations of vectors and matrices to represent multivariable Spatio-temporal PMU data. Tensors (e.g., multi-way arrays) provide a natural representation for such data [36], [37]. Tensors have found applications in many areas, such as traffic predictions [37], social network predictions [38], wireless network predictions [39], and visual data completion [40]. Using tensors, the assumption of measurement independence can be relaxed. Analyzing all modes of data and capturing correlations among multiple modes will lead to more accurate event detection and operation. The patterns of events that occur in the power system are revealed by the PMU data. For instance, the event involving loss of generation/load in one of the regions can be seen as transients in the frequency graph, whereas the ap-change position of transformers can be manifested as transients in voltage magnitude rather than frequency or voltage angle [41]. These events may be originating from the power system itself or due to cyber-attacks. In either case, voltage and current transients may have adverse effects on the health and reliability of power equipment. Thus, the timely detection of such events would help in implementing appropriate control actions in order toto maintain the resiliency, reliability, and security of the power system.

*b) Dimensionality reduction for streaming PMU data*

Authors in [15] have proposed an online reduction method for stream synchrophasor data based on the principal component analysis (PCA). The synchrophasor data dimension is reduced by finding a correlation between features such as voltage, frequency, and current, then representing it with their principal components. The proposed approach has been tested on real PMU data generated by SEL412 and GE n60 PMUs deployed on a Real-Time Digital Power System Simulator of a 4-bus and 3-generator system. Another PCA-based linear approach has been proposed in [16]. In this approach, a PCA-based method has been used to reduce the dimensionality of the data and early detect anomaly events. Events are detected by computing the error between a small set of PMU measurements, which are used as a base vector, and the predicted measurements from other PMUs. A large error between the actual and predicted measurements indicates a change in the system. Based on the obtained results from these two studies, PCA-based approaches provide promising techniques for reducing the dimensionality of online PMU data. However, the authors did not provide more details on how the proposed PCA-based approach can detect anomaly events since no analysis comparison was provided between their methods and the existing techniques.

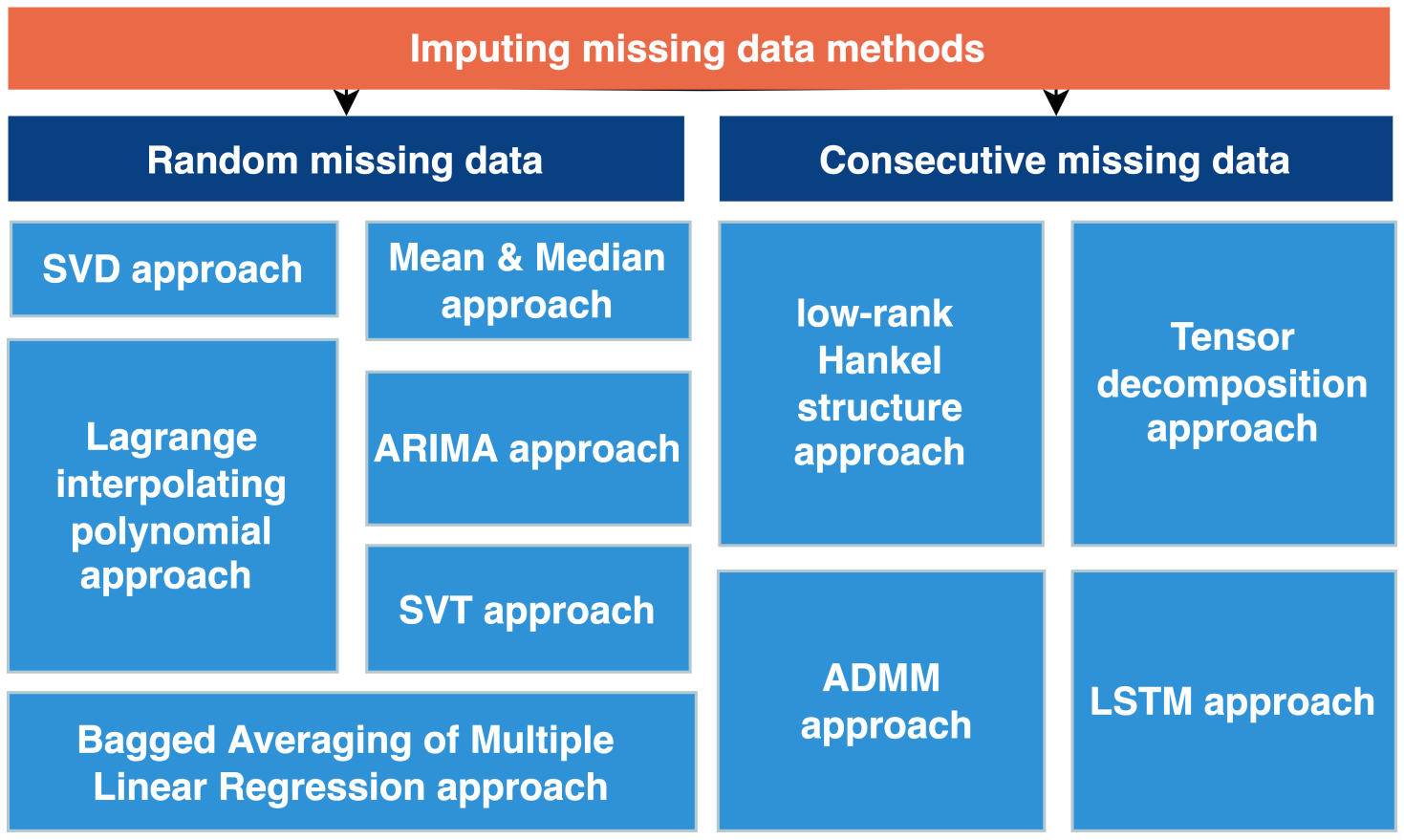

*Figure 5. Taxonomy of imputing PMU data approaches*

### c) *Imputing missing PMU data*

PMU data suffer from data quality issues such as data redundancy, bad, flawed, or missing data. Therefore, it is crucial to implement effective processes and tools to detect and deal with these data quality issues [42]. Thus, data quality is required for many power system applications to maintain the resilience of the power system. However, the quality of PMU data can be affected by errors that can occur at any stage of processing, from sensing the physical quantity to be measured to the display /analysis [43].

The PMU data frame exchanged between the PMU and the PDC can reveal some significant information about the quality of data. Based on the IEEE C37.118 standards, the data frame contains the real-time data measured or calculated by the PMU/PDC. The message includes an identification header, message length, the source ID of the message, a time stamp, and detailed status information regarding the data and its source and quality. Thus, if there are two data frames with the same identification header and the same timestamp, then it can be regarded as duplicated data.

While aggregating the data from several PMUs, the PDC may attempt to fill the unavailable data by providing a filler and setting the STAT flag in the data frame either to 10 or 11. This may indicate that the reported data is invalid. The missing data can be due to network dropout, database storage errors, or missing data from the PMU itself. It can also be detected from the PMU data frame when the STAT flag reports a value of 01.  Though getting these data quality flags is necessary, but not sufficient for identifying all the bad data.

Missing data is typically handled in two ways: deletion processing and imputation processing [44]. Deletion processing is indeed effective in eliminating missing observations in many cases as well as being easy to implement. However, some studies have found that it can severely lower the performance of subsequent models and create bias in experiments [45]. Imputation, in contrast to deletion, maintains the length of a time series by estimating missing values. Statistical interpolation is one of the most straightforward imputation methods. Various imputing PMU data are provided in Figure 5. The data gaps are filled using either mean, median, or mode of existing observations [46]. These techniques consist of recovering the missing values by computing the mean and median of the existing variables along with conserving the distribution of the data. The main drawback of these techniques is that they reduce the variance in the data and they may also delete some important frequency events such as transient or impulse. In [47], the authors, proposed a dynamic regression model comprising a regression and an autoregressive integrated moving average model (ARIMA). Using the linear relationship between neighbor and correlated stations, this model adjusts residuals by fitting an ARIMA structure. However, ARIMA is not sufficient because it needs to be combined with additional approaches such as the Kaltman filter to recover the missing data. Additionally, authors in [48]  have proposed an approach called Bagged Averaging of Multiple Linear Regression based on a bootstrapping aggregation technique in machine learning. It consists of combining multiple linear regression models into one framework to complete the missing data. It resolves the issues of the mean and ARIMA approaches however it is inefficient in dealing with PMU data which includes nonlinear relation between the frequency measurement data. In [22], the authors proposed a recovery approach designed specifically for state estimate application. This approach produces satisfactory results with a low rate of reported data, roughly five reports per second. However, a typical PMU reports the phasor information to the control center with a frequency of 60 samples per second. Thus, this approach is inefficient in dealing with and fulfilling the missing data with a high rate of reported data. Authors of [50] used Lagrange interpolating polynomial method to recover the lost PMU data. The authors assume that the PMU network uses an optical network with a loss of maximum of three continuous packets. Authors in [51] propose a Singular Value Decomposition (SVD) method that is based on projecting and lifting a subspace to impute missing data. In [52], the authors applied the Singular Value Thresholding (SVT) method to recover missing data, and it proved to be effective for up to 30% of the missing PMU. Nonetheless, these approaches have certain limitations. SVT-based methods, for instance, have high computational complexity and are only suitable for recovering PMU data in an offline setting. The tracking and estimation performances of the low-rank matrices completion method will be impaired during transients, which occur during disturbances. The online low-rank matrix completion methods should start with PMU measurements that do not have missing index values. Additionally, and even though these approaches have shown prominent results with random missing data, they are inefficient with consecutive missing data [45]. Alternatively, other approaches considered imputing consecutive missing data. In [53], the authors presented a method that could address consecutive data loss across multiple PMU channels based on a low-rank Hankel structure. In [54], the authors proposed the alternating direction multiplier method (ADMM), which also results in good results for missing data in multiple PMUs. In [55], the authors describes how to recover missing PMU data by employing tensor decomposition. In [45], the authors proposed a deep learning theory through long short-term memory (LSTM) and instance-based transfer learning through TrAdaBoost. Similarly, the authors proposed a Transferred LSTM-based iterative estimation model [56] for imputing consecutive missing data. Based on the analysis of the existing approaches it turns out that ensemble learning produces satisfactory results in terms of fulfilling the missing PMU data without losing any valuable features or changing the data distribution.

*2) ML categories*

*a) Supervised learning*

Several authors have proposed supervised function-based models for detecting anomalies using PMU data [28]-[57]. Most of these approaches were trained on data including input features such as voltage, current, phase angle, and fault location as output. These supervised-based learning models can be classified into classification and regression models, which in turn can be further categorized into function-based, decision tree-based, Bayesian and ensemble models. It is important to note that some of these models are only suitable for the offline application while others are considered online models as they are capable of handling streaming data and providing responses in near-real-time.

(1) Function-based models

Function-based models include models that rely on activation functions, such as Neural Network, or kernels, such as Support Vector Machine (SVM), to build a mapping function between the inputs and the output. Authors in [28] proposed a back propagation-based neural network (BPN) estimate fault location in distribution networks. Here, fault current was selected as a key feature to train the NN model. A Levenberg-Marquardt algorithm (also known as damped lease square) is applied to BPN for faster convergence. Then, the BNN model was deployed to run on the DIgSILENT Power Factory 13.2. Similarly, a feed-forward NN (FNN) based approach is proposed in [58]. Here, fault voltages and fault currents are selected as two features to train the model. A sigmoid activation function was used to normalize the data. Their results showed a detection error of less than 3%. Another NN-based approach was proposed in [59] to estimate fault distances from substation(s). The selected input features include: three-phase voltage, current, fault conditions, and active power gathered from substation(s). This approach was trained on different fault locations, resistances, and loads. The approach was tested on an IEEE 34-bus system and yielded promising results, even under dynamic changes in network topology. Additionally, this approach showed more tolerance to noise. Although high accuracy was reported for NN-based approaches in the above-mentioned studies, the training time required for NN is longer and does not suit dynamic or real-time environments. On the contrary, the SVM-based approach is faster and relatively accurate, even for larger-size data. However, it requires careful selection of appropriate kernel types and hyper-parameters.

In [60], the authors proposed a convolutional neural network (CNN) based approach using bus voltages. This method has been trained and tested on IEEE 39-bus and IEEE 68- bus systems under uncertain conditions for system observability and measurement quality. Their results show that CNN can localize the faulted line even in low visibility (7% of buses) conditions. Another Recurrent Neural network (RNN) has been proposed in [61] to deal with the electricity stealth issue. Specifically, a deep autoencoder has been coupled with a long-short-term-memory (LSTM)-based sequence-to-sequence (seq2seq) structure to capture false data injected in the AMI network.

A KNN based approach for detecting faults in a photovoltaic (PV) system is proposed in [62]. This approach has been trained and tested on data generated from a developed PV model. The reported results show a classification accuracy of 98.70% with an error value ranging between 0.61% and 6.5%. Another KNN based approach algorithm was applied to classify three-phase faults (3LG), voltage oscillation, and voltage sag scenarios in [63]. However, the model accuracy was not provided. In [11], the authors proposed a distributed IDS in a Multi-Layer Network Architecture of Smart Grids. In their approach, the IDSs were deployed at three levels of the smart grid network, Home Area Network, Neighborhood Area Network, and Wide Area Network. IDSs were based on the Support Vector Machine (SVM) and Artificial Immune system (IAS) where each classifier was trained and tested on the NSL_KDD dataset [12]. The obtained results illustrated that SVM outperformed AIS algorithms in terms of False Positive Rate (FPR) and False Negative Rate (FNR). However, the proposed classifiers are not suitable for streaming data since they require loading the entire dataset into memory to process it. In addition, the used dataset does not include any data related to the power system faults.

(2) Decision Tree-based models

Several online decision tree-based approaches have been proposed to deal with streaming PMU data and detect the eventual anomalies. For instance, authors [64] proposed a Hoeffding Tree (HT) combined with two concept drift detectors: drift detection method (DDM) and Adaptive sliding windows (ADWIN) for building a fast decision tree that is adaptable to changes. This model has been trained on a synthetic dataset where different attack scenarios were simulated. The dataset contains normal and anomaly events, and physical and cyber events. The physical event includes power system faults while the cyber event includes trip command injection, 1LG Fault replay, and Replay Disabled attacks. The model HT+ADWIN+DDM has been trained and tested on this heterogeneous dataset and reported a classification accuracy greater than 98% for binary classification.

In [19], the authors developed a Transfer HT with ADWIN (THTA) for detecting anomaly events in the streaming PMU data. HT has been trained on normal and anomaly events while ADWIN has been included at the leaf of the HT to make the model adaptable to the concept drift, whether gradual or abrupt. Similarly, authors in [65] have proposed a Hoeffding Adaptive Tree (HAT) based approach for detecting events from continuous streams of PMU data within computational boundaries of memory and processing time. The authors have conducted three experiments to generate a training dataset of binary classes. In the first one, synchrophasor data with three-phase faults has been generated and load fluctuations have been injected by changing the True Power (P) and Reactive Power (Q) at a regular interval to evaluate the ability of the HAT model in adapting to the concept drift. In the second one, two classes of events have been generated: fault and normal classes. The fault class includes the SLG faults while the normal class includes normal power system operations without changing the load conditions. In the third experiment, non-adaptive HT with a fixed size has been evaluated on the previously generated dataset to illustrate the impact of the tree size on the classification accuracy. Based on the reported results, HT showed the concept drift compared to traditional Decision Trees (DTs) J48 and REPTree. However, in these two studies, the duration of the physical fault events was not considered while generating the training synchrophasor dataset. Some faults have to be monitored for a certain time to identify their natures. For instance, a frequency deviation of 0.15 to 1.0 Hz is considered an inter-area oscillation only if the fluctuation lasts for 60 seconds [66]. Otherwise, it is considered a normal power system operation. In addition, some physical events are time-sensitive and if they are not early detected then they may lead to cascaded outages. Thus, it is necessary to include the event duration as a feature in the training dataset.

(3) Bayesian-based models

Authors in [67] proposed a Bayesian-based approach for detecting false injected data. Here, the authors proposed a dynamic Bayesian game-theoretic approach for detecting FDI attacks with incomplete information with a 98% detection rate. Other authors in [68] proposed a Naïve Bayed model along with RF, DT, and Logistic regression for detecting various cyber anomalies. The reported results showed satisfactory results in terms of detecting attacks in an offline and online network.

(4) Ensemble models

Authors in [69] proposed an ensemble approach based on the Active learning-based extreme gradient Boosting (AL-XGBoost). This approach was developed to detect FDI attacks in a cyber-physical energy system. The average detection accuracy obtained with the AL-XGBoost method is 0.9784 and 0.9845 for IEEE 57- and 118-bus test systems, which is higher than SVM and KNN. Authors in [70] proposed an AdaBoost combined with a Genetic Algorithm (GA) and Deep Neural Network (DNN) to detect electricity theft attacks. The model has shown superiority over SVM, ANN, and RF based on accuracy, and true positive and false-positive rates. In [71] authors discuss a fault line identification and localization approach using RF and decision tree classifiers. Here, the models were trained on an IEEE 68 Bus system. The generated data consists of several fault types including a Three-phase short circuit (TP), Line to ground (LG). The reported experiment results show that classification accuracy of 91%. Another RF-based approach was proposed in [72]. Here, the model was trained on three-phase current and voltage data, validated using an IEEE-34 system, and achieved an accuracy above 90%.

Another online ensemble approach based Intrusion Detection System (IDS) has been proposed in [13]. The proposed IDS has been deployed at three levels of the Advanced Metering Infrastructure (AMI): Smart meter, AMI headend, and Data concentrator. In their study, seven data stream mining algorithms have been selected: AccuracyUpdatedEnsemble, ActiveClassifier, LeveragingBag, LimAttClassifier, OzaBagAdwin, OzaBagASHT, and SingleClassifierDrift. These algorithms have been trained and tested on the KDD Cup 1999 dataset [14] and a performance analysis comparison has been conducted based on several metrics including Accuracy, the Kappa, the false positive rate (FPR), and false-negative rate (FNR). Based on the obtained results, ActiveClassifier and SingleClassifierDrift reported satisfactory results in terms of memory and time so they are suitable candidates for Smart Meters which have a limited computational resources. LeveraginBag provided higher accuracy and lower false positive rate but it is demanding in terms of memory. Thus, it is an appropriate candidate for the data concentrator which possesses a moderate amount of memory. For the data concentrator, which requires a response time of a fraction of a second, ActiveClassifier can be used since it processes the data quickly with modest accuracy, FPR, and FNR. However, the dataset used for training these models is not based on power system communication scenarios. Instead, it is based on the communication scenarios in computer networks. Thus, the reported results do not reflect the actual output of the studied data stream classifiers in the power system context.

#### *b) Unsupervised learning*

Authors in [73] proposed an unsupervised learning model for real-time event classification and fault localization in synchrophasor data. Their methodology relies on three processes. The first process focuses on removing bad data from collected PMU measurements using the Maximum Likelihood Estimation (MLE) approach. In the second process, the events are classified using a combination of Density-based spatial clustering of applications with noise (DBSCAN), and logic rules were generated using a physics-based decision tree (PDT) method. This PDT method uses parameters such as active power, reactive power, and fault event types. The third process reports localizing events in real-time using a graph theory. Finally, a score metric is computed using Shannon entropy, and descriptive statistical parameters (e.g., standard deviation, range, mean difference, and crest factor). Three case studies have been considered using metrics such as precision and recall and their reported result show that their proposed data cleansing approach outperforms Chebyshev and K-means methods, with a 95% precision. Additionally, the average run-time taken for their classification algorithm is around 0.09s for a typical window size of 30 samples involving five PMU sensors.
Authors in [29] proposed another unsupervised K-mean-based model for detecting DoS attacks, including UDP and ICMP flooding attacks. The proposed model was able to cluster the normal and abnormal behaviors using unlabeled data, but the model's performances were not provided.

#### *c) Semi-supervised and Reinforcement learning*

Authors in [74] proposed a semi-supervised and deep representation learning for detecting anomalies in WAMS including Short-circuit fault, Remote tripping command injection attack, and Data injection attack.
Authors in [75] proposed a multiagent RL-based approach for detecting the simple and coordinated FDI attacks on the distributed control layer in a dc microgrid. Similarly, authors in [76] modeled the anomaly detection problem as a partially observable Markov decision process (POMDP) problem and developed a universal online detection algorithm based on RL for detecting FDI, DoS, and jamming attacks. The model was evaluated in an IEEE 14-bus system and showed a precision above 99%.

#### *d) Hybrid approaches*

Several hybrid methods combining conventional and ML approaches were proposed to detect anomalies in PMU data. For instance, authors in [77] proposed a wavelet transform and Support Vector Machine (SVM) to locate faults in transmission lines, and can be described in three stages. In the first stage, voltage and current values emitted by a transmitter were used to locate the fault; the second phase feeds a multi-class SVM model to the training based on selected influential features, and classification of fault location is done using a regression approach. Here, the fault classification error is below 1% for all fault types and specifically 0.26% for SLG, 0.74% for LLG, 0.20% for LL, and 0.39% for LLLG. Authors in [78] have proposed an event location estimation (ELE) algorithm for the wide-area monitoring system for PMU data. Their approach relies on clustering and wavelet analysis to detect and localize events in real-time. In this work, the network is initially divided into several clusters, where each cluster is defined as an electrical zone using K-means. Next, a wavelet-based event detection approach is used to detect and localize event occurrences by tracking any large (e.g. event magnitudes) disturbance levels. Once the event is detected, its magnitude is defined using a Modified Wavelet Energy value, and its location is estimated at each electrical zone. The authors implemented the ELE approach in real-world PMU-setting containing 32 dynamic events with an excellent localizing accuracy value. It is important to note that the authors did not consider data quality issues in the PMU measurements. Some probable causes for data quality issues could be irregular sampling or data rate, bandwidth challenges, and time synchronization errors.

In [79], the authors discuss a wavelet decomposition technique combined with fuzzy logic to identify both faulty line(s) and their locations in a multi-terminal high voltage direct current network. In their paper, wavelet coefficients of both positive and negative currents were initially computed and then fed to a fuzzy logic-based voting system to identify the faulty line(s). Once the line is identified, a traveling wave-based algorithm is used to determine the exact fault location using the Daubechies wavelets. A discrete wavelet transform (DWT) combined with SVM for fault detection in distribution networks has been proposed in [80]. Here, features are extracted using SVM and DT and then optimized using a GA. Their model performance was evaluated on two active distribution networks (e.g., IEEE 13-Bus and IEEE 34-Bus systems), and the authors claim that their model outperforms the probabilistic neural network (PNN). A summary of all these anomaly detection approaches along with their advantages and limitations is provided in Table 1.

*Table 1. Anomaly detection approaches for PMU data*

| Category | Approach | Fault types | Anomalous event type | Advantages | Limitations |
|---|---|---|---|---|---|
| Conventional | Traveling wave [57] | Physical | Single-phase grounding fault | High tolerance to the excessive resistance, load variance, reflection, grounding resistance, refraction of the traveling wave or series capacitor bank | The accuracy of the model relies heavily on capacitance and line inductance |
| | Impedance-based [35], [59] | | Unbalanced faults | Simple and easy to implement | The accuracy can be affected in the case of a grounded fault, where the fault resistances can reach higher values. |
| Machine learning | Back propagation-based neural network (BPN) [60] | Physical | 1 phase, 2 phase, and 3 phases faults | The model is fast to converge and requires few features | The selected features are insufficient to capture various anomaly signatures. |
| | KNN [71] | Physical | Open circuit faults, line-line (LL) faults, partial shading with and with-out bypass diode faults and partial shading with inverted bypass diode faults in real-time | Fault location accuracy reaches 98.70% with an error between 0.61% and 6.5%. | The study considered the PV data only |
| | Feed-forward NN (FNN) [67] | Physical | Shunt faults | The detection error is less than 3%. High tolerance to the fault resistance, fault type, and fault location. | The fault location was not considered and they are unsuitable for real-time applications |
| | NN [68] | Physical | Ground fault, short circuit | High tolerance to noise. | |
| | CNN [69] | Physical | TP, LG, double line to ground (DLG) and line to line (LL) faults | Optimal localization estimation even under low visibility (7% of buses) | |
| | MLE + DBSCAN [81] | Physical | Three-phase fault, P load increased, Q load increased | The proposed data cleansing approach outperforms Chevyshev and K-means and achieve a precision of 95%. Less than 0.9 s to classify events for a typical window size of 30 sample data. | |
| | RF+ DT [79] | Physical | Three-phase fault, line to ground (LG), line-to-line to ground, line-to-line (LL), loss of line | Fault location detection accuracy is 91% | |
| | RF [80] | Physical | Single-phase to ground faults | Fault location detection accuracy is 90.96% in distribution systems | |
| | HT+ADWIN+DDM [72] | Cyber and physical | Symmetric, unsymmetrical faults, and trip command injection attack | Classification accuracy is greater than 94% for multiclass and greater than 98% for binary class. | Fault location and duration were not considered. This model is not suitable for real-time network |
| | HAT [73] | Cyber and physical | Three-phase faults, SLG | Optimal adaptability to the concept drift events | |
| | OzaBag and Active classifier [25] | Cyber | DoS, unauthorized access from a remote machine, to local super user (root) privileges, surveillance, and probing. | ActiveClassifier and SingleClassDrift algorithms reported satisfactory results in terms of time processing and memory consumption | |
| | RL deep Q-network [83] | Cyber and physical | FDI attack | Effective in detecting simple and coordinated FDI attacks | |
| | SVM+AIS [58] | Cyber | DoS attack and unauthorized access from a remote machine | The detection accuracy is high | Processing time is high and it is not suitable for streaming data. The fault duration was not considered |
| | POMPD [84] | Cyber and physical | FDI, DoS, and jamming attacks | Model efficient in detecting various attacks with a 99% precision | Model in unsuitable for real-time deployment. |
| | LSTM [70] | Cyber and physical | FDI attack and electricity theft | The model outperforms the Naïve Bayes, SVM, and ARIMA in terms of detection rate and false alarm with an improvement of 4−21% and 4−13%, respectively | |
| | NB, RF, DT, LR [76] | Cyber | Scanning attack | Results showed that these models are efficient in detecting SCADA cyberattacks in real-time. | Neither attack location nor duration were considered. |
| | Dynamic Bayesian game-theoretic approach [75] | Physical | Load measurement | The model is effective in detecting anomalies with incomplete information and with a detection rate of 98% | |
| | AL-XGBoost [77] | Physical | FDI attack and cascading failure | The model reaches a detection accuracy of 99% and outperforms SVM and KNN | |
| | Semi-supervised Deep representation learning [82] | Cyber and Physical | Short-circuit fault, remote tripping command injection attack, and FDI attack | The model outperforms the supervised model in terms of the true positive rate | Model suffers from a high false-positive rate compared to supervised ML models |
| | AdaBoost with GA and DNN [78] | Physical | FDI attack | Model is superior to SVM, RF, and NN in terms of detecting electrical theft with 94.8% of true positive rate | The attack model was not provided. |
| | K-means [61] | Cyber | DoS attack | Model is efficient in handling unlabeled data | Model's performances were not provided. |
| Hybrid | SVM+ Wavelet transform [85] | Physical | SLG, LLG, LL | The fault classification error is below 1% for all fault types. | The fault duration was not considered and the model is not suitable for streaming power system data. |
| | K-mean + Wavelet transform + ELE [86] | Physical | 32 dynamic events including generator and line trips | Fault location accuracy attain 100% | The fault duration was not considered and the processing time is high, which make them inappropriate for streaming data. |
| | Fuzzy logic + Wavelet transform [87] | Physical | Negative to ground and the positive to ground faults | The error between the actual fault location and the predicted one is low then 0.002% | |
| | SVM+ Discrete wavelet transform [88] | Physical | High impedance fault | Fault location accuracy is 98.27% for IEEE 13-Bus and 98.29% for the IEEE 34-Bus test systems | |

### C. Evaluation performance metrics

#### 1) Supervised Offline ML approaches

For evaluating the supervised offline ML approaches performances in the matter of detecting the cyber and physical anomalies, several metrics can be considered including the accuracy, Kappa, Probability of detection, miss detection, and fault alarms. The accuracy evaluates the models in terms of distinguishing accurately the anomalous and normal events. It is expressed as:

$$Accuracy = \frac{TP+TN}{TP+TN+FP+FN} \quad (1)$$

Where $TP$ is the true positive and corresponds to the number of cyber/physical anomalies correctly classified as anomalies. $TN$ is the true negative and corresponds to the number of normal events classified as normal. $FP$ is the false positive and corresponds to the number of normal events misclassified as anomalous events. $FN$ is the false negative corresponding to the number of anomalous events that are incorrectly classified as normal events. The $TP, TN, FP, FN$ can be derived from the confusion matrix. The accuracy may provide overly prediction results with imbalanced classes. The Kappa metrics can be used to address this issue and control the instances that may have been correctly classified by chance. It is expressed as:

$$Kappa = \frac{(Accuracy - random\ accuracy\ )}{(1 - random\ accuracy)} \quad (2)$$

Where $Accuracy$ is given by Equation (1), and $random\ accuracy$ is defined as:

$$random\ accuracy = \frac{(TN+FP)*(TN+FN)+(FN+TP)*(FP+TP)}{Total^2} \quad (3)$$

Here $Total$ is $TP + TN + FN + FP$. The probability of detection $(Pd)$ is the number of anomalous events divided by the total number of anomalous events in the dataset. It is given by:

$$Pd = \frac{\mathrm{TP}}{\mathrm{TP+FN}} \quad (4)$$

The probability of false alarm $(Pfa)$ is the number of events that are incorrectly classified as anomalies, while in fact, they are normal. It is expressed as:

$$Pfa = \frac{\mathrm{FP}}{\mathrm{FP+FN}} \quad (5)$$

The probability of miss detection $(Pmd)$ is the number of events that are incorrectly classified as normal, while in fact, they are anomalies. It is given by:

$$Pmd = \frac{\mathrm{FN}}{\mathrm{TP+FN}} \quad (6)$$

#### 2) Unsupervised ML approaches

For evaluating the unsupervised and regression ML models, the Mean Average Error (MAE) and Mean Squared Error (MSE) can be selected. The MAE is the average of the absolute differences between the actual and predicted value, and it is given by:

$$MAE = \frac{1}{n}\sum_{i=1}^{n} |\hat{y} - y| \quad (7)$$

where y^ is the predicted value, y is the actual value, and n is the number of instances or cases. Unlike MAE, MSE has the benefit of penalizing for significant errors because it averages the squared differences between the actual value and the predicted one, expressed as:

$$MAE = \frac{1}{n}\sum_{i=1}^{n} (\hat{y} - y)^2 \quad (8)$$

#### 3) Supervised Online ML approaches

With streaming PMU data, it is challenging to store the entire dataset and split it into training, validation, and testing data to evaluate the real-time ML models. Therefore, other evaluation techniques are considered for evaluating these models' performances, particularly, Holdout, interleaved test-then-train, and prequential. In Holdout, a section of the incoming data is used to train the model and small test cases were used to compute the performance. In interleaved test-and-train, each instance of the data stream is used for testing and training the model. Prequential is similar to interleaved test-and-train and uses a sliding window or a decaying factor.

## V. Research directions

Loading and processing an open-ended source of fast and huge volumes of PMU data is challenging in terms of affording the required computational resources. Additionally, the conventional ML approaches, such as decision trees, used for extracting insights are no longer relevant since they require loading and scanning the entire dataset [81]. Thus, two important criteria have to be satisfied by any ML approach when it comes to dealing with streaming PMU data: building the model with a limited dataset as it is challenging to store the entire streaming PMU measurements into memory, and the model must adapt itself to concept drift when the data distribution changes gradually or abruptly over time, which could be due to load changes [82]. Otherwise, the model built in the past will no longer be consistent with the data received in the present and its performance will decrease. Developing a classifier that meets these requirements is challenging, so approximate solutions can be considered with an associated error to be minimized [83], [84].

Due to the high potential of unexpected events that could occur in the distribution of the streaming PMU data, it is important to deal with these unexpected events and help the ML model in adapting and classifying these unknown events, this event is known as concept drift. A concept drifts when the law underlying the data changes gradually or abruptly. Thus, the model built in the past is no longer consistent with the data received in the present. Concept drifts could be due to different reasons including noise, the influence of the environment, variation in characteristics not considered in the model, and seasonal alterations. A convenient treatment of concept drift means that recent data that conflict with past assumptions should prioritize the construction of the model. But, at the same time, assumptions based on old data that are still consistent should be preserved. Several approaches have been proposed for tackling such events. These approaches include the drift detection method (DDM), Early Drift Detection Methods (EDDM), Linear Four Rate (LFR), Just in time (JIT), and ensemble methods [85], [86], [87]. DDM and EDDM don't require to store of the data but they are susceptible to false alarms. LFR [85] is constant in space complexity and can deal with imbalance classes as it uses different error types separately, but it still suffers from the labeling cost. JIT [87] does not require labeled data and can detect abrupt change but it is inefficient with the gradual drift. Ensemble methods [25] are effective in detecting the recurring concepts, but they use a large batch which makes them ineffective in identifying precisely the change location. ADWIN turns out to be an optimal concept drift detector since it is better for change localization and uses a dynamic window that adapts its size according to the change rate observed in the data within the window. When the data is stationary, the window size dynamically increases, and it decreases when a change is detected. Thus, it is important to include the ADWIN model into online ML approaches to deal with the potential concept drifts.

## VI. Conclusion

In this paper, a comprehensive overview of the cyber and physical anomalies was presented, along with an analysis of their impact on the availability and integrity of the WAMS. A critical review of current anomaly detection approaches was conducted. These approaches were divided into conventional, machine learning, and hybrid approaches, and then their strengths and limitations were assessed. Lastly, it provided insight into a variety of statistical and machine learning-based approaches to deal with missing PMU data and concept drifts.